\documentclass[runningheads]{llncs}
\usepackage[T1]{fontenc}

\usepackage{newtxtext}       %
\usepackage[varvw]{newtxmath}       

\usepackage{booktabs}   
\usepackage{makecell}   
\usepackage{pifont}     
\usepackage{colortbl}   
\usepackage{multirow}
\usepackage{graphicx}
\usepackage{threeparttable}

\definecolor{grayrow}{gray}{0.9}
\begin{document}
\title{EmboTeam: Asymmetric State Grounding for Heterogeneous Multi-Robot Collaboration}
%
%
\author{Haishan Zeng\inst{1,2} \and
Mengna Wang\inst{1,2} \and
Peng Li\inst{1,2,3}\thanks{Corresponding author.}}
\titlerunning{EmboTeam: Asymmetric State Grounding}
%
\institute{Institute of Software, Chinese Academy of Sciences\and
University of Chinese Academy of Sciences\and
Nanjing Institute of Software Technology\\
}
\maketitle

\begin{abstract}

Heterogeneous multi-robot collaboration over long horizons requires
both global task planning and robust coordination during execution.
Existing planners using large language models (LLMs) can generate task decompositions or symbolic plans, but often execute them as rigid action sequences, making them fragile to local state mismatches and synchronization failures.
We propose EmboTeam, a neuro-symbolic framework for asymmetric state
grounding. An LLM generates structured state expectations, which are verified and coordinated through symbolic search using Planning Domain Definition Language (PDDL) and then grounded into a team-level parallel behavior tree. Robot-specific subtrees
perform precondition checking, skill execution, effect validation, local
replanning, and blackboard-based synchronization. We further introduce
MACE-THOR, containing 42 heterogeneous multi-robot tasks across 8 indoor
layouts. EmboTeam improves task success rate from 12\% to 55\% and goal
condition recall from 32\% to 75\% over LaMMA-P, demonstrating the
effectiveness of synchronized reactive grounding for collaboration over long horizons.

\end{abstract}
\section{Introduction}
\label{sec:introduction}

Enabling heterogeneous robot teams to accomplish collaborative tasks over
long horizons is an important step toward embodied intelligence~\cite{li2025largelanguagemodelsmultirobot,rizk2019cooperative}.
Unlike single-robot planning, such collaboration additionally requires
capability-aware task allocation, temporal dependency preservation, shared
resource coordination, and synchronization of intermediate states.
The key challenge is therefore not only whether individual actions are
executable, but whether the team can maintain consistent collaborative
states over long horizons.

\begin{figure}[!t]
        \centering
        \includegraphics[width=\linewidth]{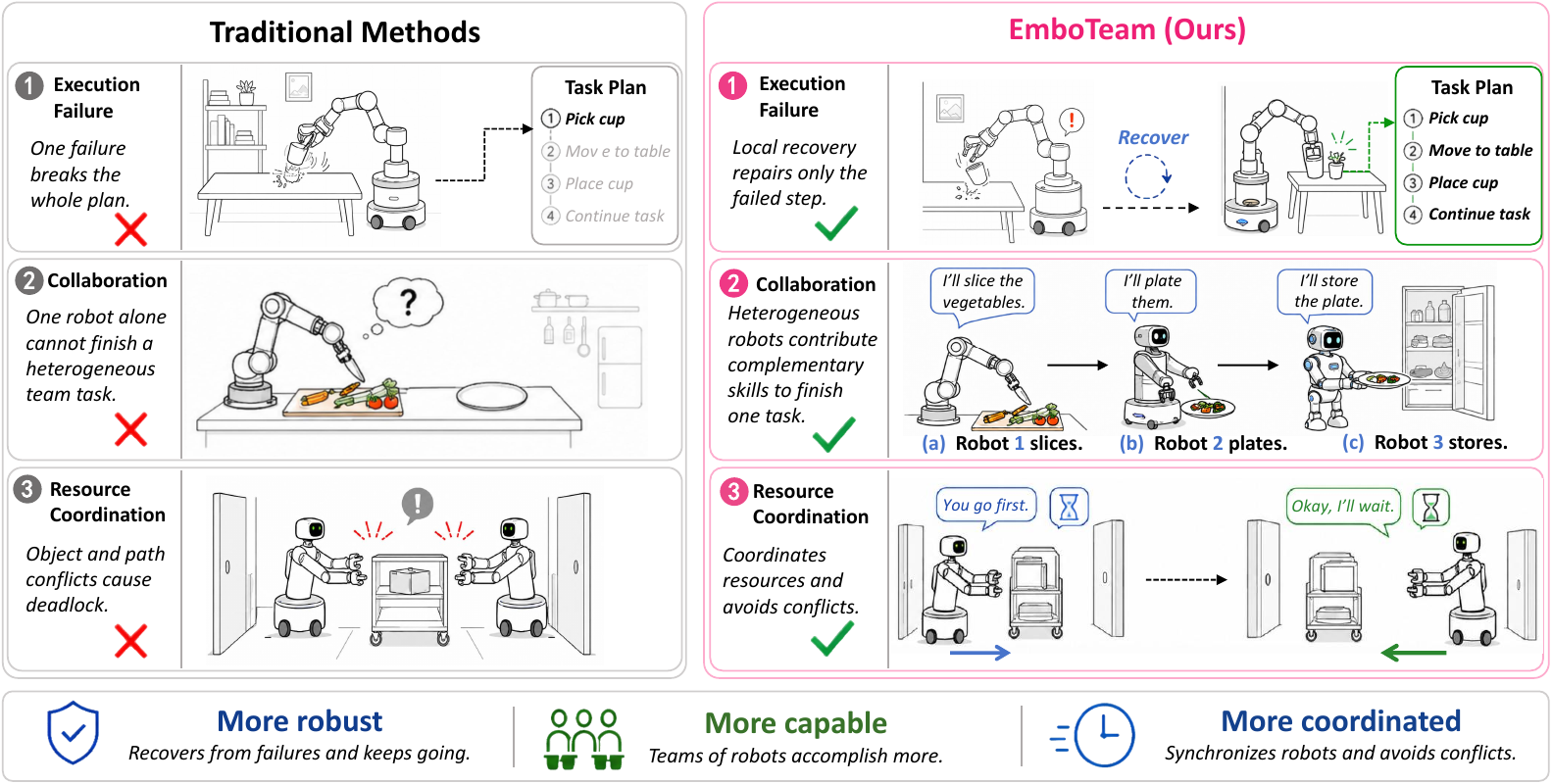}
        \vspace{-2.3em}
        \caption{\textbf{Motivation.} Traditional plan-to-action methods often execute rigid action sequences, making them brittle to local failures, heterogeneous capability requirements, and resource conflicts. EmboTeam grounds state expectations into synchronized reactive execution, enabling local recovery, complementary robot collaboration, and resource coordination.}
        \label{fig:motivation}
        \vspace{-1.5em}
\end{figure}

Recent embodied planners using large language models (LLMs) can translate natural-language
instructions into task decompositions, role assignments, symbolic plans,
or action sequences~\cite{saycan2022arxiv,kannan2023smart,song2023llmplanner}, while several approaches further integrate
the Planning Domain Definition Language (PDDL)~\cite{Fox_2003} for logical verification~\cite{liu2023llm+,silver2024generalized,zhang2025lamma}. However, execution is often
treated as a linear or loosely coupled action sequence, making long-horizon
collaboration fragile when local preconditions, intermediate effects, or
synchronization dependencies are violated. This reflects a structural
mismatch between globally structured symbolic planning and local dynamic
execution. As illustrated in Fig.~\ref{fig:motivation}, traditional plan-to-action execution is brittle when local failures, heterogeneous capability requirements, or shared resource conflicts arise during multi-robot execution. Instead of treating the generated plan as a rigid action sequence, EmboTeam explicitly transforms high-level planning outputs into state expectations and grounds them into synchronized reactive execution.

We argue that this fragility stems from a structural mismatch between the high-level cognitive planning space and the low-level embodied execution space. High-level planners usually operate over abstract, complete, and stable symbolic states, which are suitable for reasoning about task order, resource constraints, and inter-robot dependencies. In contrast, robots execute actions in a local, dynamic, and potentially inconsistent execution space, where action failures, delayed state updates, resource occupancy, and unmet effects may occur. Directly incorporating all local execution mismatches into a unified global planning space would lead to severe state-space expansion. On the other hand, ignoring these mismatches and relying on open-loop execution makes long-horizon collaboration brittle. Therefore, long-horizon embodied collaboration requires a decoupled formulation: the high-level cognitive module should generate globally consistent state expectations, while the low-level execution module should verify, ground, and recover these expectations at runtime.

Based on this observation, we propose EmboTeam, an asymmetric state
grounding framework that decouples heterogeneous multi-robot collaboration
into a global belief space and a reactive grounding space. In the global
belief space, an LLM generates structured state expectations and PDDL-based
symbolic search verifies and coordinates them. In the reactive grounding
space, these expectations are grounded into robot-specific behavior-tree
subtrees with runtime state checking, local replanning, and explicit
blackboard synchronization. This design preserves global collaboration
semantics while allowing local execution discrepancies to be handled
reactively. EmboTeam focuses on the high-level collaboration layer and is complementary to upstream perception modules and low-level vision language action (VLA) controllers.

Our main contributions are summarized as follows:

\begin{itemize}
    \item An \textbf{asymmetric state grounding formulation} is proposed for heterogeneous multi-robot collaboration. It decouples collaboration into state expectation generation in a high-level global belief space and state expectation grounding in a low-level reactive execution space, avoiding the need to directly incorporate all local execution mismatches into the global planning space.

    \item \textbf{EmboTeam}, a predict-and-verify neuro-symbolic collaboration
    framework, is developed. An LLM generates structured state expectations,
    which are verified and coordinated through PDDL-based symbolic search and
    then grounded into a team-level parallel behavior tree. Runtime
    precondition checking, effect validation, local replanning, and shared
    blackboard synchronization continuously maintain these expectations
    during execution.

    \item \textbf{MACE-THOR} is constructed for systematic evaluation. MACE-THOR contains 42 long-horizon heterogeneous multi-robot tasks across 8 indoor layouts, covering both Parallel-Independent and Temporal-Dependent collaboration scenarios. Experiments show that EmboTeam significantly improves task success rate and goal condition recall, demonstrating the effectiveness of state expectation grounding in complex collaborative tasks.
\end{itemize}
\section{Related Work}
\label{sec:related_work}


Solving complex tasks that require sequential multi-step decision-making over extended horizons is a central problem in embodied intelligence. Traditional approaches mainly include hierarchical task networks, PDDL-based symbolic planning systems, Monte Carlo tree search, and reinforcement learning~\cite{browne2012survey,georgievski2014overview,jiang2019task}. These methods provide structured mechanisms for task decomposition, symbolic reasoning, state-space search, or policy learning. However, when applied to large-scale embodied environments, they often face limitations in computational efficiency, scalability, and generalization. In particular, long-horizon collaboration among heterogeneous robots introduces additional challenges beyond single-agent planning, such as capability-aware task allocation, resource competition, temporal dependency preservation, and inter-robot state synchronization. Recent vision-language-action models and embodied foundation models have achieved promising progress in single-robot perception and short-horizon manipulation~\cite{brohan2023rt2visionlanguageactionmodelstransfer,driess2023palme}. Nevertheless, such models still require a higher-level coordination mechanism when deployed in multi-robot scenarios, where the system must maintain global task intent and intermediate states over long temporal horizons. 


The rapid development of large language models has introduced new paradigms for embodied task planning. Existing studies leverage the natural language understanding and commonsense reasoning abilities of LLMs to transform high-level instructions into task decompositions, role assignments, action sequences, or structured planning representations~\cite{saycan2022arxiv,kannan2023smart,mandi2024roco,song2023llmplanner}. A particularly active direction is to combine LLMs with PDDL-based planning~\cite{liu2023llm+}, where natural language instructions are mapped into formal planning problems and solved or verified by classical planners~\cite{liu2023llm+,mahdavi2024leveraging,silver2024generalized,zhang2025lamma}. Other works explore multi-agent planning architectures that improve task efficiency through role specialization and collaborative reasoning~\cite{kannan2023smart,mandi2024roco,singh2024twostep}.

Despite these advances, many existing methods still focus primarily on plan
generation, while treating execution as a linear action sequence or a loosely
coordinated set of agent scripts. Such execution becomes fragile in long-horizon
multi-robot collaboration when local preconditions, intermediate effects, or
synchronization dependencies are violated. A plan that is valid in the symbolic planning space may still fail during
execution due to local precondition mismatch, missing synchronization events,
resource conflicts, or unmet action effects. Moreover, inter-robot communication in many existing systems remains implicit or simplistic, making the overall performance highly dependent on the LLM's ability to maintain long-range dependencies. In contrast, EmboTeam does not directly regard LLM outputs as final executable plans. Instead, LLMs are used to generate structured state expectations, which are further verified through symbolic planning and grounded into a team-level parallel behavior tree for runtime checking, replanning, and synchronization.


Neuro-symbolic approaches combine the semantic flexibility of neural
models with symbolic reasoning for precondition, effect, and goal
verification. However, symbolic validity alone does not guarantee robust
embodied execution, where local conditions, action outcomes, and
inter-robot dependencies may change at runtime. This motivates reactive
execution mechanisms that continuously maintain symbolic collaboration
semantics.

Behavior Trees (BTs) have been widely adopted in robotics as a modular and
reactive execution structure, providing explicit control-flow semantics,
hierarchical task organization, and localized failure handling
~\cite{Colledanchise_2018,iovino2022survey}. Recent studies further
explore automated BT grounding and the integration of language-model-based
planning with BTs for long-horizon robot execution
~\cite{cai2026cabto,suzuki2026dialogue}. However, these approaches primarily
focus on constructing or executing robot-level behavior structures, while
cross-robot intermediate-state dependencies, event-driven synchronization,
and heterogeneous collaboration are not their central formulation. In contrast, EmboTeam treats BTs not as fixed execution scripts, but as a
runtime grounding structure for globally specified state expectations.
Each expectation is instantiated as a robot-specific reactive subtree with
precondition checking, local replanning, effect validation, and explicit
wait/update synchronization through a shared blackboard. This design enables
high-level collaboration semantics to be continuously verified and maintained
during multi-robot execution.
\section{Methods}
\label{sec:methods}

\subsection{Problem Formulation: Asymmetric State Grounding}

We consider a team of heterogeneous robots
\begin{equation}
    \mathcal{R}=\{R_1,R_2,\ldots,R_N\},
\end{equation}
where each robot $R_i$ is associated with a skill set $\mathcal A({R}_i)$, such as navigation, grasping, slicing, opening, or object transportation. Given a natural language instruction $I$, an initial environment state $s_0$, and a heterogeneous robot team $R$, the objective is to generate an executable team policy that collaboratively satisfies the task goal implied by $I$.

High-level planning and embodied execution operate over asymmetric state
spaces. We therefore distinguish a global belief space $\mathcal{B}$,
where LLM-based semantic reasoning and PDDL-based symbolic search operate
over globally structured states, and a reactive grounding space
$\mathcal{E}$, where robots verify local conditions, perform local
replanning, and synchronize intermediate execution states.

EmboTeam represents the collaborative task by a set of intermediate state
expectations
\begin{equation}
\mathcal{Z}=\{z_1,z_2,\ldots,z_M\},
\end{equation}
where each $z_k$ specifies the assigned robot, required state, executable
skill, expected effect, and inter-robot synchronization relation. Rather
than explicitly modeling all execution discrepancies in $\mathcal{B}$,
these expectations are monitored and maintained in $\mathcal{E}$.

\subsection{Overview of EmboTeam}

\begin{figure}[!t]
        \centering
        \includegraphics[width=\linewidth]{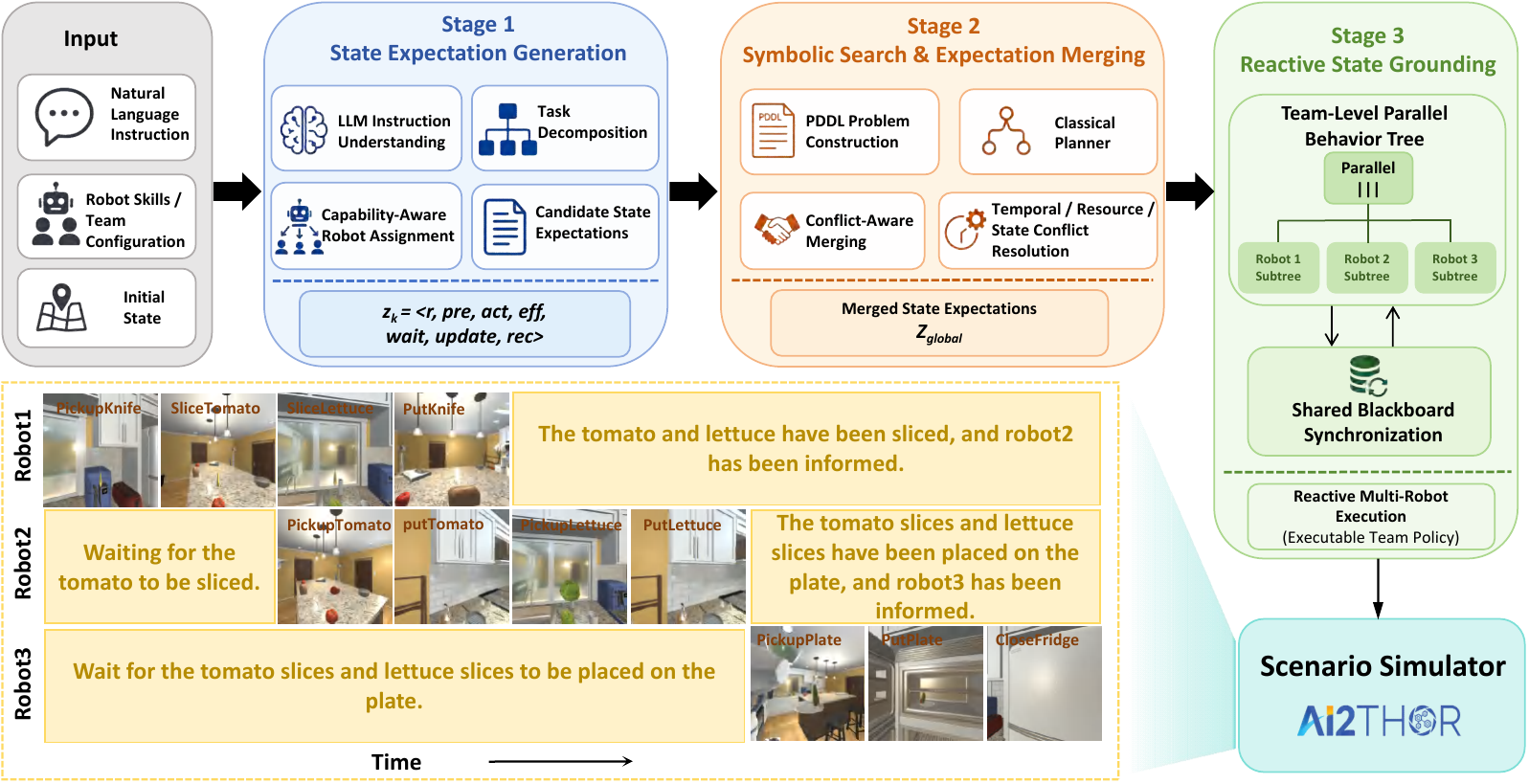}
        \vspace{-2em}
        \caption{\textbf{Overview of EmboTeam.} EmboTeam generates structured state expectations, verifies and merges them through symbolic search, and grounds them into a team-level parallel behavior tree for execution.}
        \label{fig:overview}
        \vspace{-1em}
\end{figure}

As shown in Fig.~\ref{fig:overview}, EmboTeam follows a three-stage predict-and-verify
architecture. State Expectation Generation (SEG) maps the task instruction,
robot capabilities, and initial state into structured expectations
$\mathcal{Z}$. Symbolic Search and Expectation Merging (SSEM) verifies
their symbolic feasibility with PDDL and resolves cross-expectation
temporal, resource, and state constraints, producing coordinated
expectations $\mathcal{Z}_{global}$. Reactive State Grounding (RSG)
grounds $\mathcal{Z}_{global}$ into robot-specific reactive subtrees under
a team-level Parallel node, with runtime state checking, local replanning,
and blackboard synchronization.

\subsection{State Expectation Generation}

Given $I$, $\mathcal{R}$ and $s_0$, the State Expectation Generator produces a set of structured state expectations. Each expectation is defined as
\begin{equation}
    z_k=\langle r_k, pre_k, act_k, eff_k, wait_k, update_k, rec_k\rangle,
\end{equation}
Each state expectation follows a predefined structured schema rather than
free-form natural language. Specifically, $r_k \in \mathcal{R}$ denotes
the assigned robot, $pre_k$ and $eff_k$ are sets of grounded symbolic
predicates defined over the PDDL predicate vocabulary, and
$act_k \in \mathcal{A}(r_k)$ denotes an executable skill of robot $r_k$.
The fields $wait_k$ and $update_k$ are sets of discrete synchronization
event identifiers maintained on the shared blackboard. Finally, $rec_k$
specifies a predefined local replanning strategy that is invoked when the
expected execution state cannot be restored directly.

The symbolic fields $pre_k$ and $eff_k$ describe physical world states,
such as $holding(R_2,knife)$ or $sliced(tomato)$, and are used for both
symbolic planning and runtime state verification. In contrast,
$wait_k$ and $update_k$ do not represent physical states:
$wait_k$ specifies the synchronization events required before execution,
whereas $update_k$ specifies the events published after successful effect
validation. These events are used only for execution-level coordination
through the shared blackboard. The recovery field $rec_k$ does not
represent a dynamically generated behavior-tree fragment; instead, it
invokes the predefined local replanning procedure when recovery is required.

This representation constrains the role of the LLM to generating
structured state expectations over predefined robot-skill, symbolic-predicate,
and synchronization-event vocabularies, rather than directly producing
unconstrained executable action sequences. The subsequent symbolic planning
stage verifies their feasibility, while behavior-tree grounding enables
runtime checking, replanning, and synchronization.

\subsection{Symbolic Search and Expectation Merging}

For each state expectation $z_k$, EmboTeam constructs a task-specific
PDDL problem $\mathcal{P}_k$ using its symbolic fields and the corresponding
task context. As defined in Sec.~3.3, $pre_k$ and $eff_k$ consist of grounded
symbolic predicates from the predefined predicate vocabulary and can therefore
be directly instantiated in the PDDL problem representation. Thus, no
free-form natural-language condition is directly passed to the planner.
A classical planner is then invoked as

\begin{equation}
\pi_k = Planner(\mathcal{D}, \mathcal{P}_k),
\end{equation}

Here, $\mathcal{D}$ denotes the predefined PDDL domain, $\mathcal{P}_k$ denotes the task-specific problem associated with $z_k$, and $\pi_k$ denotes the resulting feasible symbolic plan. The plan $\pi_k$ is used to verify reachability and expose symbolic dependencies for cross-expectation coordination, rather than being directly executed by the robot. The domain $\mathcal{D}$ is fixed throughout all experiments and is not
generated or modified by the LLM.

Given the fixed domain, EmboTeam instantiates $\mathcal{P}_k$ for each
state expectation using the task-relevant objects, current structured state,
and expected symbolic conditions. The involved entities are declared as
problem objects, the available symbolic state is encoded in the initial-state
specification, and $eff_k$ defines the local planning goal. The resulting
problem is therefore task-specific, while the underlying domain and action
semantics remain shared across tasks. 

However, independently generated subplans may not form a globally consistent multi-robot plan. They may contain temporal conflicts, resource conflicts, or state conflicts. Temporal conflicts occur when a subtask is scheduled before its required synchronization event is produced. Resource conflicts occur when multiple robots attempt to manipulate or occupy the same object or workspace simultaneously. State conflicts occur when the effect of one subplan fails to satisfy, or even contradicts, the precondition of another.

Although each $\pi_k$ verifies the symbolic feasibility of an individual
state expectation, independently verified expectations may still exhibit
temporal, resource, or state conflicts.

\begin{equation}
    \mathcal{Z}_{global}=Merge\left(
\{(z_k,\pi_k)\}_{k=1}^{M}
\right).
\end{equation}

The resulting $\mathcal{Z}_{global}$ is a partially ordered set of
coordinated state expectations with explicit wait/update dependencies.
The symbolic plans $\pi_k$ provide feasibility and coordination evidence
but are not directly compiled into the behavior tree~\cite{cai2026cabto,Colledanchise_2018,suzuki2026dialogue}.

\subsection{Reactive State Grounding via Behavior Trees}

The Reactive State Grounding module compiles the merged global state expectations $\mathcal{Z}_{global}$ into an executable team-level behavior tree. Instead of executing the global plan as a flat action sequence, EmboTeam represents the whole multi-robot task as a parallel behavior tree:
\begin{equation}
    \mathcal{T}_{team}=Parallel(\mathcal{T}_{R_1},\mathcal{T}_{R_2},\ldots,\mathcal{T}_{R_N}),
\end{equation}
where $\mathcal{T}_{R_i}$ denotes the robot-specific subtree assigned to robot $R_i$. Each robot subtree contains the state expectations assigned to that robot, while cross-robot dependencies are maintained through explicit wait/update synchronization events. Thus, the behavior tree is not merely a local wrapper for individual actions, but the executable structure of the entire multi-robot task.

For each state expectation $z_k \in \mathcal{Z}_{global}$, EmboTeam generates a reactive behavior subtree:

\begin{equation}
\small
\begin{aligned}
\mathcal{T}(z_k)=\operatorname{Seq}\big(
&\operatorname{Wait}(wait_k),
\operatorname{Fallback}(\operatorname{Check}(pre_k),\operatorname{Recover}(rec_k)),\\
&\operatorname{Execute}(act_k),
\operatorname{Validate}(eff_k),
\operatorname{Update}(update_k)
\big).
\end{aligned}
\label{eq:bt_compile}
\end{equation}

Here, the Wait blocks execution until all required synchronization events are available on the shared blackboard. The Check verifies whether the current local execution state satisfies the symbolic precondition. If the check fails, the Recover node invokes the predefined local
replanning procedure to regenerate an executable local action sequence
under the current symbolic state. The replanning is restricted to the
corresponding robot subtree and state expectation, avoiding unnecessary
global replanning whenever the mismatch can be resolved locally. The Execute calls the corresponding robot skill. The Validate checks whether the expected postcondition is achieved after execution. Finally, the Update writes completion events to the blackboard to activate downstream subtasks.

The compilation follows a hierarchical structure. At the top level, $\mathcal{T}_{team}$ coordinates all robots through a parallel node. At the robot level, each $\mathcal{T}_{R_i}$ encodes the ordered or partially ordered expectations assigned to robot $R_i$. At the expectation level, each $\mathcal{T}(z_k)$ provides runtime condition checking, local replanning, effect validation, and event update. This hierarchy allows EmboTeam to preserve global collaboration semantics while giving each robot the ability to react to local execution mismatches.

This compilation step is central to EmboTeam. It transforms high-level state expectations into executable structures that specify not only what each robot should do, but also when it is allowed to act, what conditions must be verified, how local mismatches should be recovered, and how completion states should be synchronized across robots. As a result, high-level symbolic plans are grounded into reactive multi-robot execution policies rather than brittle open-loop action sequences.

\begin{figure}[!t]
        \centering
        \includegraphics[width=\linewidth]{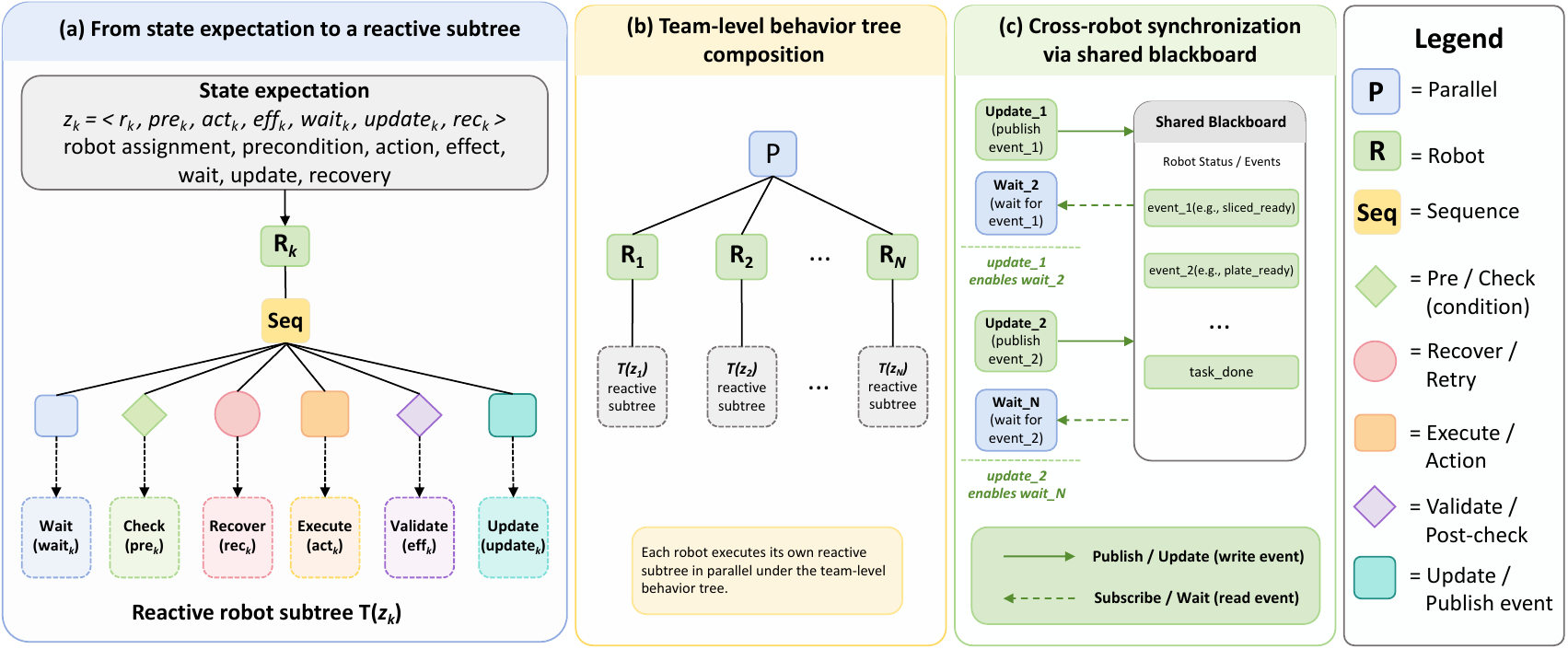}
        \caption{\textbf{Reactive grounding and synchronization in the team-level behavior tree.} (a) Each state expectation is compiled into a robot-specific reactive subtree with wait, check, recover, execute, validate, and update nodes. (b) Robot subtrees are organized under a team-level parallel behavior tree. (c) Cross-robot dependencies are maintained through wait/update events over a shared blackboard.}
        \label{fig:synchronization}
    %
    %
    %
\end{figure}

\subsection{Blackboard Synchronization and Discrepancy Resolution}

Long-horizon multi-robot failures often arise not from a single invalid action,
but from inconsistent intermediate states among robots. For example, a downstream
robot may start execution before an upstream robot has completed a prerequisite,
or a state assumed by the global plan may not actually hold during execution.
EmboTeam addresses this issue through a shared blackboard that explicitly stores
synchronization events and intermediate state updates. Fig.~\ref{fig:synchronization}
illustrates this mechanism from reactive subtree compilation to team-level blackboard synchronization.

When a robot completes a state expectation, the corresponding Update node writes an event to the blackboard. Other robots whose expectations depend on this event are blocked by their Wait nodes until the event becomes available. If a robot detects that its local execution state does not match the
high-level expectation, it first invokes local replanning within the
corresponding behavior subtree. If local replanning fails to restore an
executable state, or a required synchronization event cannot be produced
within the predefined timeout, the robot writes a global failure flag to
the shared blackboard. All active robot subtrees continuously monitor this
flag. Once it is set, pending Wait nodes terminate with Failure rather than
remaining in the Running state, and the failure is propagated to the
team-level Parallel node, which terminates the current collaborative
execution. This timeout-and-abort mechanism prevents downstream robots from waiting
indefinitely for synchronization events that can no longer be produced,
thereby avoiding deadlock caused by upstream failures.

This mechanism localizes recoverable execution-time discrepancies within
individual robot subtrees, while unrecoverable failures are explicitly
propagated through the shared blackboard. It prevents indefinite blocking
of dependent robots and reduces the risk of cascading failures in
long-horizon collaboration, without repeatedly invoking global replanning.
\section{Experiments}
\label{sec:experiments}

\begin{table}[t]
\centering
\caption{\textbf{Benchmark Statistics.} Comparison of task scale and complexity across multi-robot planning benchmarks.}
\label{tab:benchmark_statistics}
\vspace{-2mm}
\resizebox{\linewidth}{!}{
\setlength{\tabcolsep}{4.5pt}
\begin{tabular}{lccccccc}
\toprule
\textbf{Benchmark}
& \textbf{Tasks}
& \textbf{Layouts}
& \textbf{Robots}
& \textbf{Avg. goals}
& \textbf{Avg. subtasks}
& \textbf{Avg. objects}
& \textbf{Dependency depth} \\
\midrule
SMART-LLM~\cite{kannan2023smart}
& 36 & 7 & 1--4 & 1.58 & 5.50 & 2.19 & 4.28 \\
MAT-THOR~\cite{zhang2025lamma}
& 70 & 7 & 2--4 & 1.76 & 6.63 & 2.47 & 4.94 \\
MACE-THOR-PI
& 21 & 6 & 2--3 & 2.14 & 10.19 & 3.19 & 7.71 \\
MACE-THOR-TD
& 21 & 8 & 2--4 & 3.38 & 19.00 & 4.24 & 16.67 \\
\textbf{MACE-THOR}
& 42 & 8 & 2--4 & 2.76 & 14.60 & 3.71 & 12.19 \\
\bottomrule
\end{tabular}
}
\vspace{-2mm}
\end{table}

\subsection{Benchmark and Task Categories}

To evaluate long-horizon heterogeneous multi-robot collaboration~\cite{kang2026mars},
we construct MACE-THOR on top of AI2-THOR~\cite{kolve2017ai2}. Inspired by MAP-THOR~\cite{nayak2024mapthor}
and MAT-THOR~\cite{zhang2025lamma}, MACE-THOR provides 42 household tasks across 8 indoor layouts for evaluating heterogeneous robot team collaboration, as summarized in Table~\ref{tab:benchmark_statistics}.
Each task includes a natural language instruction, heterogeneous robots, an initial
environment state, and goal conditions. The benchmark supports teams of 2 to 4
robots and covers capability-aware task allocation, temporal dependency preservation,
and synchronized execution. 

The tasks are divided into two categories. \textbf{Parallel-Independent tasks} can be decomposed into multiple subtasks with weak or no execution dependencies, allowing robots to work in parallel. These tasks mainly evaluate task decomposition, robot assignment, and parallel execution. \textbf{Temporal-Dependent tasks} contain strict predecessor constraints, where downstream subtasks can only be executed after upstream robots produce required intermediate states. These tasks are more challenging and directly evaluate whether a method can preserve state expectations and synchronization relations during execution.

\subsection{Evaluation Metrics and Baselines}

We adopt four evaluation metrics. Success Rate (SR) measures the proportion of tasks that are fully completed. Goal Condition Recall (GCR) measures the proportion of goal conditions that are achieved, reflecting partial task completion. Executability (Exec) measures whether the generated action sequence can be physically executed by the simulator, regardless of whether the final task goal is fully satisfied. Execution-to-Success Conversion (ESC) measures how effectively executable task executions are converted into full task success. 

We compare EmboTeam with representative LLM-based multi-agent planning baselines, including CoT prompting, SMART-LLM~\cite{kannan2023smart} and LaMMA-P~\cite{zhang2025lamma}. These methods are closely related to language-driven task decomposition, multi-agent role assignment, and embodied multi-agent planning. For a fair comparison, all methods are evaluated in the same AI2-THOR environments and under the same task definitions. For the main comparison, SMART-LLM, LaMMA-P, and EmboTeam are evaluated with GPT-4o; we additionally report EmboTeam variants using Llama-3.1 and Claude-3.5.

\begin{table}[!t]
\caption{\textbf{Comparative Evaluation.} Results on Parallel-Independent and Temporal-Dependent tasks.}
\label{tab:comparison}
\vspace{-0.5em}
\centering
\scriptsize
\setlength{\tabcolsep}{2.5pt}
\renewcommand{\arraystretch}{0.95}

\begin{tabular*}{\linewidth}{@{\extracolsep{\fill}}lcccccccc@{}}
\toprule
\multirow{2}{*}{\textbf{Methods}}
 & \multicolumn{4}{c}{\textbf{Parallel-Independent}} 
 & \multicolumn{4}{c}{\textbf{Temporal-Dependent}} \\
\cmidrule(lr){2-5} \cmidrule(lr){6-9}
& \textbf{SR $\uparrow$} & \textbf{GCR $\uparrow$} & \textbf{Exec $\uparrow$} & \textbf{ESC $\uparrow$}
& \textbf{SR $\uparrow$} & \textbf{GCR $\uparrow$} & \textbf{Exec $\uparrow$} & \textbf{ESC $\uparrow$} \\
\midrule

CoT (GPT-4o)
& $0.05$ & $0.10$ & $0.68$ & $0.07$
& $0.00$ & $0.00$ & $0.00$ & $0.00$ \\

SMART-LLM (GPT-4o)
& $0.10$ & $0.18$ & $\mathbf{0.83}$ & $0.12$
& $0.00$ & $0.00$ & $0.00$ & $0.00$ \\

LaMMA-P (GPT-4o)
& $0.14$ & $0.40$ & $0.81$ & $0.17$
& $0.10$ & $0.26$ & $\mathbf{0.83}$ & $0.12$ \\

Ours (Llama-3.1)
& $0.52$ & $0.60$ & $0.82$ & $0.63$
& $\mathbf{0.38}$ & $0.50$ & $0.76$ & $0.50$ \\

Ours (Claude-3.5)
& $0.57$ & $0.70$ & $0.77$ & $0.74$
& $0.33$ & $0.55$ & $0.57$ & $\mathbf{0.58}$ \\

Ours (GPT-4o)
& $\mathbf{0.71}$ & $\mathbf{0.88}$ & $0.76$ & $\mathbf{0.93}$
& $\mathbf{0.38}$ & $\mathbf{0.62}$ & $0.71$ & $0.54$ \\

\bottomrule
\end{tabular*}
\vspace{-1em}
\end{table}

\subsection{Results and Analysis}

\textbf{Main results.} Table~\ref{tab:comparison} reports the comparative results on MACE-THOR. EmboTeam consistently improves SR and GCR over SMART-LLM and LaMMA-P across both task categories. Averaged over all tasks, EmboTeam increases the SR from 12\% to 55\% and GCR from 32\% to 75\% compared with LaMMA-P. Although some baselines achieve high Exec, their low SR, GCR, and ESC indicate that physically executable actions do not necessarily satisfy long-horizon collaborative goals. In contrast, EmboTeam achieves substantially higher ESC, showing that its executable behaviors are more effectively converted into successful task completion. This improvement comes from grounding intermediate state expectations into synchronized reactive execution rather than executing open-loop action sequences.

\begin{figure}[!b]
        \centering
        \includegraphics[width=\linewidth]{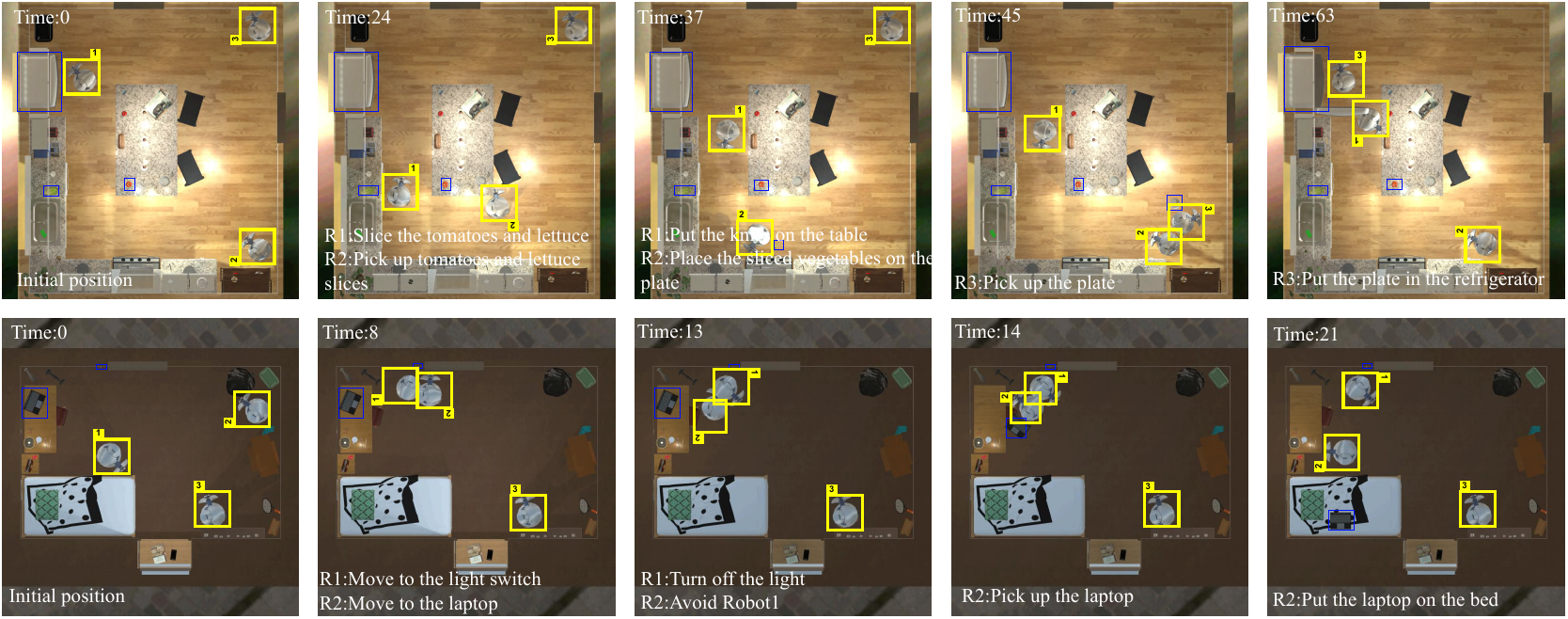}
        \vspace{-2em}
        \caption{\textbf{Execution keyframes in AI2-THOR.} Examples show synchronized collaboration in a Temporal-Dependent kitchen task and concurrent execution in a Parallel-Independent task.}
        \label{fig:keyframes}
        \vspace{-2em}
\end{figure}

\textbf{Qualitative analysis.}
Fig.~\ref{fig:keyframes} shows qualitative AI2-THOR executions for both
Temporal-Dependent and Parallel-Independent tasks. On Parallel-Independent
tasks, EmboTeam benefits from capability-aware decomposition and team-level
parallel behavior tree execution. On Temporal-Dependent tasks, the advantage
mainly comes from explicit wait/update synchronization and runtime validation
of state expectations. The performance gap is especially clear in Temporal-Dependent tasks, where open-loop execution baselines often fail to preserve the
required intermediate states.

\textbf{Mechanism analysis.}
EmboTeam improves robustness by grounding intermediate state expectations
into robot-specific behavior tree subtrees. Each robot waits for required events,
checks local preconditions, validates effects, and updates the shared blackboard,
which is especially important for Temporal-Dependent tasks where downstream
robots must not execute before upstream states are satisfied. This mechanism also explains the higher ESC of EmboTeam, since executable actions are continuously checked, validated, and synchronized before contributing to downstream task completion.

\begin{table}[!t]
\caption{\textbf{Ablation Study.} Results under the asymmetric state grounding formulation. SEG, SSEM, and RSG denote state expectation generation, symbolic search and expectation merging, and reactive state grounding.}
\label{tab:ablation}
\vspace{-0.5em}
\centering
\scriptsize
\setlength{\tabcolsep}{2.5pt}

\begin{tabular*}{\linewidth}{@{\extracolsep{\fill}}lcccccccc@{}}
\toprule
\multirow{2}{*}{\textbf{Methods}}
 & \multicolumn{4}{c}{\textbf{Parallel-Independent}} 
 & \multicolumn{4}{c}{\textbf{Temporal-Dependent}} \\
\cmidrule(lr){2-5} \cmidrule(lr){6-9}
& \textbf{SR $\uparrow$} & \textbf{GCR $\uparrow$} & \textbf{Exec $\uparrow$} & \textbf{ESC $\uparrow$}
& \textbf{SR $\uparrow$} & \textbf{GCR $\uparrow$} & \textbf{Exec $\uparrow$} & \textbf{ESC $\uparrow$} \\
\midrule

Ours (w/o SEG \& SSEM)
& $0.43$ & $0.65$ & $0.55$ & $0.78$
& $0.14$ & $0.41$ & $0.53$ & $0.26$ \\

Ours (w/o SSEM)
& $0.43$ & $0.65$ & $0.82$ & $0.52$
& $0.19$ & $0.22$ & $0.51$ & $0.37$ \\

Ours (w/o RSG)
& $0.14$ & $0.40$ & $0.81$ & $0.17$
& $0.10$ & $0.26$ & $0.83$ & $0.12$\\

Ours (w/o sync)
& $0.48$ & $0.68$ & $0.83$ & $0.58$
& $0.10$ & $0.21$ & $0.56$ & $0.18$\\

Ours (w/o recovery)
& $0.33$ & $0.43$ & $\mathbf{0.97}$ & $0.34$
& $0.19$ & $0.31$ & $\mathbf{0.95}$ & $0.20$\\

Ours
& $\mathbf{0.71}$ & $\mathbf{0.88}$ & $0.76$ & $\mathbf{0.93}$
& $\mathbf{0.38}$ & $\mathbf{0.62}$ & $0.71$ & $\mathbf{0.54}$ \\

\bottomrule
\end{tabular*}
\vspace{-3em}
\end{table}

\textbf{Ablation study.} To evaluate how each part of asymmetric state grounding contributes to execution robustness, we ablate state expectation generation (SEG), symbolic search and expectation merging (SSEM), and reactive state grounding (RSG). Table~\ref{tab:ablation} reports the results. Removing SEG and SSEM weakens task decomposition, robot assignment, and symbolic verification, leading to lower SR and Exec. Removing only SSEM preserves partial executability but hurts Temporal-Dependent tasks, showing that independently generated expectations must be merged under temporal, resource, and state constraints. Removing RSG causes a large drop in SR, GCR, and ESC, even when Exec remains high. This indicates that without runtime checking, local replanning, and explicit synchronization, physically executable actions are not reliably converted into successful task completion. These results show that high-level plans become reliable only when their state expectations are grounded into a synchronized and recoverable team-level execution structure.


\section{Conclusion}

We presented EmboTeam, an asymmetric state grounding framework for
long-horizon heterogeneous multi-robot collaboration. EmboTeam represents
language-specified tasks as structured state expectations, verifies and
coordinates them through PDDL-based symbolic search, and grounds them into
a team-level parallel behavior tree with runtime checking, local
replanning, and blackboard synchronization. We further introduced
MACE-THOR, containing 42 heterogeneous multi-robot tasks across 8 indoor
layouts. Experiments demonstrate substantial improvements over
representative LLM-based planning methods, particularly on tasks requiring
long dependency chains and intermediate-state synchronization. Future work
will explore tighter integration with learned visual grounding, VLA
controllers, and physical multi-robot systems.

\bibliographystyle{splncs04}
\bibliography{references}
\end{document}